\documentclass[runningheads]{llncs}
\usepackage[T1]{fontenc}
\usepackage{graphicx,verbatim}
\usepackage{amsmath} 
\usepackage{amssymb} 
\usepackage{makecell} 
\usepackage{booktabs} 
\usepackage{tabularx} 
\usepackage{multirow} 
\usepackage{algorithm} 
\usepackage{algorithmic} 
\usepackage{hyperref} 
\usepackage{url} 
\usepackage{colortbl} 
\usepackage{xcolor} 
\usepackage{subfigure} 
\definecolor{best}{HTML}{C5E1A5}
\definecolor{second}{HTML}{D9EAC6}
\definecolor{third}{HTML}{EFF6E8}
\hypersetup{
    colorlinks=true,
    citecolor=blue,
    linkcolor=blue,
    urlcolor=blue
}
\begin{document}
\title{Two-Stage Cross-Domain Cervical Abnormality Screening with Cytopathological Image Synthesis and Knowledge Distillation}
\titlerunning{Cross-Domain Cell Detection.}
%

\author{
Jincheng Li\inst{1}\textsuperscript{*} \and
Yuzhi He\inst{2}\textsuperscript{*} \and
Yihui Zhan\inst{1} \and
Xinmei Zhang\inst{1} \and
Yifei Sun\inst{3} \and
Zelin Liu\inst{4} \and
Lichi Zhang\inst{4} \and
Minye Shao\inst{5} \and
Lili Zhao\inst{1}\textsuperscript{\dag}
}
\authorrunning{J. Li et al.}
\institute{School of Artificial Intelligence and Computer Science, Nantong University, Nantong 226019, China \and
School of Telecommunications Engineering, Xidian University, Xi'an 710126, China \and
College of Computer Science and Technology, Zhejiang University, Hangzhou 310058, China \and
School of Biomedical Engineering, Shanghai Jiao Tong University, Shanghai 200030, China \and
Department of Computer Science, Durham University, Durham, UK\\[4pt]
{\small $^*$These authors contributed equally to this work.}\\
{\small $^\dagger$Corresponding author: \email{ylzh@ntu.edu.cn}}}
  
\maketitle              
\begin{abstract}
Cross-domain diagnosis remains a major challenge in cervical cell pathology due to pronounced domain shifts across institutions and the subtle visual differences among disease stages, which jointly impair model generalization. To address these issues, this paper proposes a two-stage framework for cross-domain cervical cell detection. In the first stage, we propose the Spatially-Continuous Unpaired Neural Schrödinger Bridge (SC-UNSB), which constructs a synthetic intermediate domain to mitigate cross-domain distribution shifts by modeling image translation as an entropy-regularized optimal transport process. In the second stage, we propose a dual-level feature alignment strategy within a knowledge distillation, which progressively aligns shallow structural features and deep semantic representations to facilitate the transfer of domain-invariant knowledge from the source to the target model. Experimental results demonstrate that the proposed method effectively mitigates domain shift and category ambiguity, improving the cross-domain detection performance. Code is available at \url{https://github.com/ZhanYiHui06/MICCAI2026}

\keywords{Cross-domain cervical cell  \and Schrödinger bridge \and Knowledge distillation.}

\end{abstract}

\section{Introduction}

As the fourth most common cancer globally, cervical cancer remains a leading cause of mortality among women, making early detection critical for prognosis and survival \cite{cohen2019cervical,bray2020erratum}. Recently, object detection models based on deep learning have demonstrated robust performance on datasets of cervical cytology \cite{fei2024distillation,li2025high,fei2025weakly,hu2025controllable}. However, their generalization across institutions remains limited. Variations in sample preparation workflows and staining protocols across medical institutions often lead to substantial performance degradation in cross-institutional evaluations. Moreover, cervical carcinogenesis is a continuous biological process, where visual differences between lesion stages are inherently subtle. Under cross-domain, such fine-grained distinctions become even more difficult to discern, increasing the risk of misclassification and potential clinical harm \cite{zhu2024advancing,zhu2026medeyes}.

Domain adaptation techniques have been introduced to mitigate cross-domain discrepancies. Among them, knowledge distillation provides a mechanism for transferring representations from a model trained on a source domain to a target-domain model \cite{lan2026reco,lan2025acam}. Existing distillation-based adaptation methods typically perform feature alignment at a single representation level or treat shallow and deep features independently \cite{pan2025deformable,shen2024two,tong2025uncertainty}. Such strategies may not fully account for the hierarchical nature of cytological representations, where structural information and semantic abstraction contribute differently to cross-domain generalization. Furthermore, when appearance discrepancies between domains are substantial, feature-level alignment alone may be insufficient, motivating the integration of image-level translation mechanisms.

To address these challenges, this paper proposes a two-stage framework for cross-domain cervical cell detection. Our contributions can be summarized as follows:

\begin{itemize}
    \item To mitigate severe appearance discrepancies across institutions, we propose the Spatially-Continuous Unpaired Neural Schrödinger Bridge (SC-UNSB), a spatially continuous image translation framework that constructs an intermediate domain by enforcing dense statistical normalization within the Schr\"odinger Bridge formulation. This design explicitly addresses boundary-induced artifacts in ultra-high-resolution cytopathological images.

    \item To reduce semantic misalignment under cross-domain settings, we introduce a dual-level feature alignment strategy within a knowledge distillation framework. By jointly aligning shallow structural representations and high-level semantic embeddings, the proposed method facilitates progressive transfer of domain-invariant knowledge.

    \item Extensive experiments on two large-scale cervical cytology datasets demonstrate that the proposed framework consistently improves cross-domain detection performance, achieving up to 26.9\% mAP and 45.8\% mAP50.
    
\end{itemize}

\section{Methodology}
\subsection{Spatially-Continuous Unpaired Neural Schr\"{o}dinger Bridge }

To mitigate cross-domain appearance discrepancies, we build upon the Unpaired Neural Schr\"{o}dinger Bridge (UNSB)~\cite{kim2023unpaired}, which formulates unpaired image translation as an entropy-regularized optimal transport problem. Given source and target distributions $\pi_0$ and $\pi_1$, the Schr\"{o}dinger Bridge seeks a diffusion process $\{X_t\}_{t\in[0,1]}$ by solving
\begin{equation}
    \min_{\mathbb{Q}} \mathcal{KL}(\mathbb{Q} \,\|\, \mathbb{W}) 
    \quad \text{s.t.} \quad 
    \mathbb{Q}_0 = \pi_0,\; \mathbb{Q}_1 = \pi_1,
\end{equation}

where $\mathbb{W}$ denotes the reference Wiener measure. In practice, UNSB approximates this objective via a discrete-time Markov process and adversarial learning.

While UNSB effectively aligns global distributions, directly applying it to ultra-high-resolution cytopathological images requires patch-wise translation, which introduce statistical discontinuity and boundary artifacts across neighboring regions. In standard generators, Instance Normalization (IN) computes feature statistics $\mu^{(k)}$ and $\sigma^{(k)}$ independently for each patch $x^{(k)}$:

\begin{equation}
    \text{IN}(x^{(k)}) = \gamma \frac{x^{(k)} - \mu^{(k)}}{\sigma^{(k)}} + \beta
\end{equation}

This independence, however, triggers a Boundary Drift Error: the statistical discontinuity at the patch interface $\partial \Omega_k$ causes the optimal transport manifold to become spatially fractured. Such fractures manifest as tiling artifacts, which can distort cell structures and jeopardize clinical diagnosis.

\begin{figure}[t]
\centering
\includegraphics[width=\columnwidth]{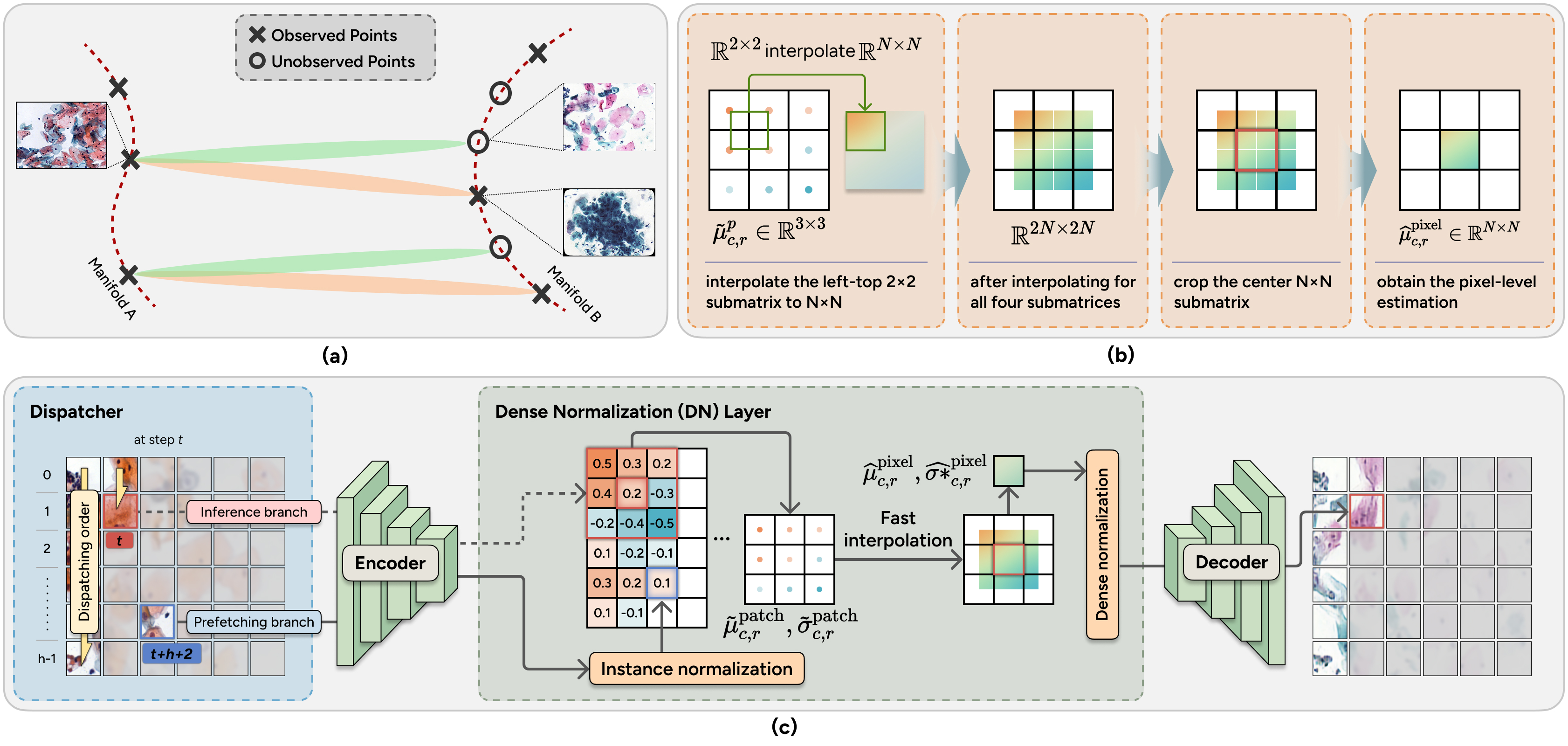}
\caption{Overview of SC-UNSB. (a) Learning process of the Schr\"odinger Bridge for entropy-regularized transport between source and target distributions. (b) Dense pixel-level moment estimation, where an $N \times N$ statistical field is interpolated from a $3 \times 3$ neighborhood of patch-level statistics. (c) Architecture of the dispatcher and the Dense Normalization (DN) layer. An ultra-high-resolution image is divided into patches. The dispatcher coordinates prefetching and inference branches, and pixel-wise mean $\hat{\mu}_{r,c}$ and variance $\hat{\sigma}_{r,c}$ are obtained via fast interpolation from neighboring patches to ensure spatially continuous normalization.}
\label{fig1}
\end{figure}

To rectify the transport path, we propose the Spatially-Continuous Unpaired Neural Schrödinger Bridge (SC-UNSB). Instead of treating statistical moments as scalars, we re-parameterize them as continuous functions of pixel coordinates $p=(u,v)$ within the generator. We use a Dense Normalization (DN) module into the UNSB generator, as illustrated in Fig.~\ref{fig1}. For a specific pixel $p$ in patch $k$, its statistics are estimated by interpolating the coarse statistics from the current patch and its neighbors $\mathcal{N}_k$. Let $\mathcal{S}_k = \{(\mu^{(n)}, \sigma^{(n)}) \mid n \in \mathcal{N}_k\}$ be the set of regional statistics. We define a bilinear interpolation operator $\Phi_{\text{interp}}$ to construct the smooth statistical fields:

\begin{equation}
    \hat{\mu}(p) = \Phi_{\text{interp}}(\{\mu^{(n)}\}_{n \in \mathcal{N}_k}, p), \quad \hat{\sigma}(p) = \Phi_{\text{interp}}(\{\sigma^{(n)}\}_{n \in \mathcal{N}_k}, p)
\end{equation}

Consequently, the generator $G_{\theta}$ is modified to be position-aware, utilizing these dense fields for normalization:

\begin{equation}
    G_{\text{SC}}(x^{(k)}, t) = \text{Conv} \left( \gamma \odot \frac{x^{(k)} - \hat{\mu}(p)}{\hat{\sigma}(p)} + \beta \right)
\end{equation}

By integrating $\hat{\mu}(p)$ and $\hat{\sigma}(p)$, we implicitly constrain the search space of the Schr\"{o}dinger Bridge to the subspace of spatially continuous functions. This ensures that the generated cell structures maintain topological consistency across patch boundaries, effectively eliminating tiling artifacts while preserving high-frequency biological details.

\subsection{Distillation-based Feature Alignment}

\begin{figure}[t]
\centering
\includegraphics[width=\columnwidth]{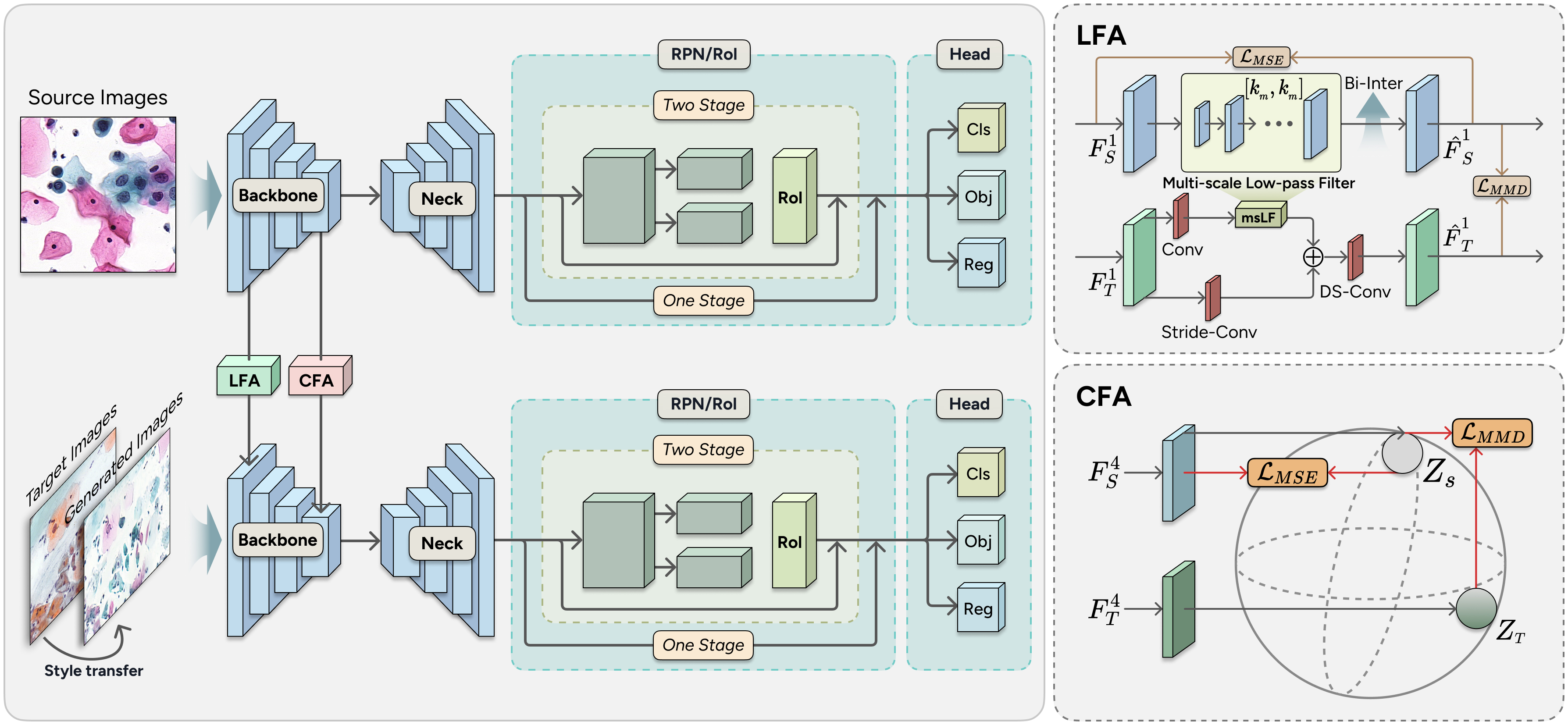}
\caption{Model of the dual-level feature alignment. The source model guides the target model via LFA in the frequency domain for structural patterns, and CFA for high-level semantics.}
\label{fig2}
\end{figure}

We propose a dual-level feature alignment strategy within a knowledge distillation framework, as illustrated in Fig.~\ref{fig2}. The proposed framework consists of two complementary components: Loose Feature Alignment (LFA), which focuses on preserving shallow structural representations at early stages, and Compact Feature Alignment (CFA), which emphasizes the transfer of high-level semantic knowledge.

To reduce the marginal distribution discrepancy between source and target domains, we adopt $\mathcal{L}_{MMD}$ as the cross-domain alignment objective. Meanwhile, a reconstruction constraint $\mathcal{L}_{MSE}$ is imposed to preserve the original source-domain representations during alignment. The overall feature alignment losses are defined as

\begin{equation}
\mathcal{L}_{MMD} = \frac{1}{B} \left\| \sum_{i=1}^{B} \phi(\hat{F}_S^i) - \sum_{j=1}^{B} \phi(\hat{F}_T^j) \right\|_2^2 , \mathcal{L}_{MSE} = \| F_S - \hat{F}_S \|_2^2
\end{equation}

where $\phi(\cdot)$ denotes the feature mapping function, $\hat{F}_S$ and $\hat{F}_T$ are the aligned source and target features, and $B$ is the batch size.

Loose Feature Alignment (LFA) operates on shallow features to preserve structural information that is less sensitive to semantic variation but vulnerable to domain shift. We transform features into the frequency domain using a multi-scale low-pass filter (MSLF). For the source model, average pooling with different kernel sizes extracts low-frequency components, followed by bilinear interpolation:

\begin{equation}
\hat{F}_{S} = MSLF(F_{S}) = \Phi(AvgPool_{k_{a} \times k_{a}}(F_{S}))
\end{equation}

where $\text{AvgPool}_{K_a \times K_a}$ denotes the average pooling function with a kernel size of $K_a \times K_a$, and $\Phi(\cdot)$ represents the bilinear interpolation operation. 

Regarding the target domain model feature $\hat{F}_T$, considering the discrepancies in shallow features between the source and target domain models, we design a learnable low-pass filter. This filter is composed of a multi-scale low-pass filter, a convolutional downsampling module, and depthwise separable convolution (DSConv). The learnable target domain feature $\hat{F}_T$ in the frequency domain can be expressed as:

\begin{equation}
\hat{F}_{T} = Conv_{3 \times 3}(Concat[DownSample_{s \times s}(F_{T}), MSLF(F_{T})])
\end{equation}

where $\text{DownSample}_{S \times S}$, $\text{Concat}$, and $\text{Conv}_{3 \times 3}$ denote the convolutional downsampling module, the feature concatenation operation, and the $3 \times 3$ convolution operation, respectively. 

After obtaining the frequency-domain features of the source and target domain models through the MSLF transformation, the loss for LFA can be calculated using the loss formula, which can be expressed as $ \mathcal{L}_{LFA} = \mathcal{L}_{MMD} + \mathcal{L}_{MSE}$

Compact Feature Alignment (CFA) aligns high-level semantic representations from the penultimate layer. Since encoder outputs may differ in dimension, a $1\times1$ convolution projects features into a unified embedding space $Z$. Alignment in this compact space promotes transfer of class-discriminative knowledge. Similarly, the loss for CFA can be expressed as $\mathcal{L}_{CFA} = \mathcal{L}_{MMD} + \mathcal{L}_{MSE}$

By maintaining the original training paradigm of the target domain model while employing a coarse-to-fine strategy—transitioning from loose to compact feature alignment—the target domain model is enabled to acquire knowledge from the source domain model, with a specific emphasis on domain-invariant features. The overall training loss can be expressed as:

\begin{equation}
\mathcal{L} = \underbrace{\mathcal{L}_{cls} + \mathcal{L}_{loc}}_{task} + \alpha \cdot (\underbrace{\mathcal{L}_{LFA} + \mathcal{L}_{CFA}}_{feature \ alignment})
\end{equation}

where $\mathcal{L}_{cls}$ and $\mathcal{L}_{loc}$ represent the classification and localization losses for the cell detection task, respectively, and $\alpha$ serves to balance the various loss components.

\begin{figure}[b!]
\centering
\includegraphics[width=\columnwidth]{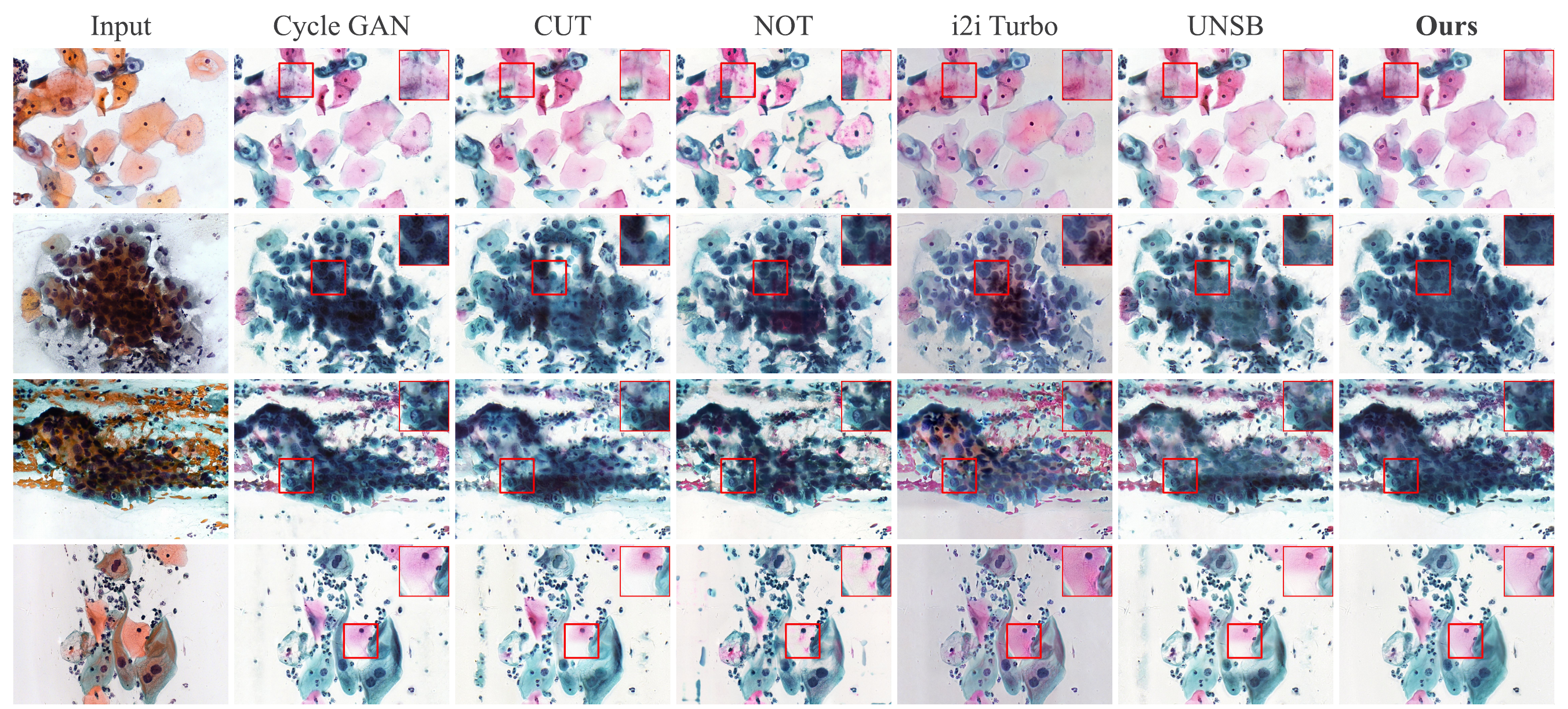}
\caption{Qualitative comparison between different generators in transferring $\mathcal{D}_s$ images into $\mathcal{D}_i$ images. Each method produces four examples in individual rows. Zoom in to check image details.}
\label{fig3}
\end{figure}

\section{Experiment}

\begin{table}[ht]
\centering
\scriptsize
\setlength{\tabcolsep}{4pt}
\caption{Comparison with other methods. The best, second best and third best results are highlighted in \colorbox{best}{dark}, \colorbox{second}{medium} and \colorbox{third}{light} green, respectively.}
\label{tab:1}
\begin{tabular}{l|ccccccc}
\toprule
\textbf{Generator} & $\mathcal{D}_s$ & CycleGAN\cite{zhu2017unpaired} & CUT\cite{park2020contrastive} & NOT\cite{korotin2022kernel} & i2i-Turbo\cite{parmar2024one} & UNSB\cite{kim2023unpaired} & \textbf{SC-UNSB} \\
\midrule
FID $\downarrow$ & 241.05 & 147.28 & \cellcolor{best}132.61 & 177.43 & 154.65 & \cellcolor{third}143.43 & \cellcolor{second}135.31 \\
KIDx100 $\downarrow$ & 10.896 & \cellcolor{third}1.831 & \cellcolor{best}1.365 & 5.104 & 2.514 & 2.411 & \cellcolor{second}1.807 \\
NIQE $\downarrow$ & 14.72 & \cellcolor{third}13.51 & 14.6 & \cellcolor{second}12.86 & 16.39 & 13.94 & \cellcolor{best}11.38 \\
HIST $\uparrow$ & 0.384 & 0.695 & \cellcolor{second}0.722 & 0.571 & 0.473 & \cellcolor{third}0.701 & \cellcolor{best}0.754 \\
\midrule
\textbf{Detector} & \multicolumn{7}{c}{\textbf{RetinaNet}} \\
\midrule
mAP $\uparrow$ & 4.6\% & 10.5\% & \cellcolor{third}14.0\% & 9.3\% & 10.3\% & \cellcolor{second}17.8\% & \cellcolor{best}20.2\% \\
mAP50 $\uparrow$ & 9.3\% & 22.4\% & \cellcolor{third}29.3\% & 18.0\% & 26.8\% & \cellcolor{second}35.4\% & \cellcolor{best}41.5\% \\
\midrule
\textbf{Detector} & \multicolumn{7}{c}{\textbf{RetinaNet $+$ CFA}} \\
\midrule
mAP $\uparrow$ & 7.8\% & 12.1\% & \cellcolor{third}16.3\% & 9.6\% & 13.9\% & \cellcolor{second}21.2\% & \cellcolor{best}22.7\% \\
mAP50 $\uparrow$ & 19.1\% & 23.7\% & \cellcolor{third}33.8\% & 20.4\% & 30.1\% & \cellcolor{second}38.7\% & \cellcolor{best}43.1\% \\
\midrule
\textbf{Detector} & \multicolumn{7}{c}{\textbf{RetinaNet $+$ LFA}} \\
\midrule
mAP $\uparrow$ & 5.1\% & 12.3\% & \cellcolor{third}19.4\% & 12.1\% & 12.4\% & \cellcolor{second}23.8\% & \cellcolor{best}24.1\% \\
mAP50 $\uparrow$ & 11.5\% & 24.0\% & \cellcolor{third}35.4\% & 18.4\% & 28.4\% & \cellcolor{second}41.1\% & \cellcolor{best}43.7\% \\
\midrule
\textbf{Detector} & \multicolumn{7}{c}{\textbf{RetinaNet $+$ LFA $+$ CFA}} \\
\midrule
mAP $\uparrow$ & 12.6\% & 18.8\% & \cellcolor{third}22.7\% & 11.3\% & 15.1\% & \cellcolor{second}24.1\% & \cellcolor{best}26.9\% \\
mAP50 $\uparrow$ & 26.6\% & 31.8\% & \cellcolor{third}40.3\% & 21.9\% & 30.9\% & \cellcolor{second}42.6\% & \cellcolor{best}45.8\% \\
\bottomrule
\end{tabular}
\end{table}

\begin{figure}
\centering
\includegraphics[width=\columnwidth]{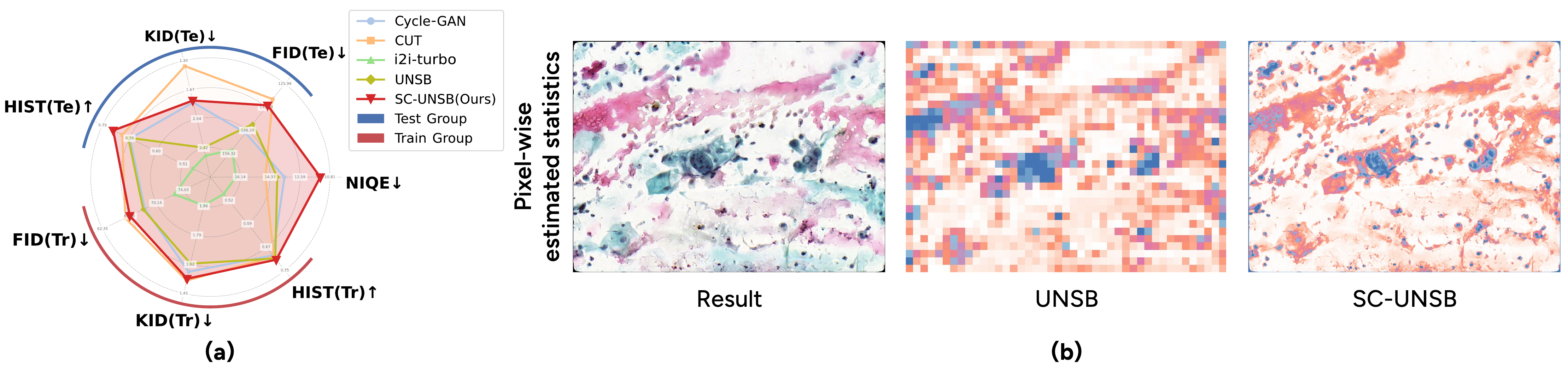}
\caption{(a) Radar chart comparison of multiple quantitative metrics; (b) Visual comparison of pixel-wise estimated statistics.}
\label{fig4}
\end{figure}

{\textbf{Datasets and Settings:}} We conducted extensive experiments using the CRIC dataset\cite{rezende2021cric}, which contains 7,839 cervical microscopy bounding boxes, as the source domain ($\mathcal{D}_s$), and the ComparisonDetector dataset\cite{liang2021comparison}, which contains 7,410 clustered cell images from Pap smear slides, as the target domain ($\mathcal{D}_t$). The ($\mathcal{D}_t$) splits are reconstructed for strict cross-domain evaluation and are released with the code. Models are trained using stochastic gradient descent with a learning rate of 0.001 and momentum 0.9. Pseudo-labels with confidence above 0.9 supervise the student model. Evaluation covers image generation quality and detection performance on the target-domain test set, which is strictly excluded from training.

For generation, we compare SC-UNSB with five representative unpaired translation methods that map $\mathcal{D}_s$ to the intermediate domain ($\mathcal{D}_i$). We leverage FID \cite{heusel2017gans} and KID \cite{chen2020reusing} scores to assess the diversity and quality of the generated images. Additionally, NIQE \cite{ho2022ultra} and HIST are  utilized to evaluate the naturalness and realism of the synthesized results. For detection, RetinaNet is trained on $\mathcal{D}_i$ and $\mathcal{D}_t$, followed by knowledge distillation. Performance is evaluated using mAP\cite{lin2014microsoft} and mAP50.

{\textbf{Generation:}} As shown in Table~\ref{tab:1}, the high FID of source images indicates a substantial domain gap. While CycleGAN and UNSB partially reduce this gap, their relatively high FID and NIQE scores indicate limited alignment in distribution statistics and perceptual quality.SC-UNSB achieves the lowest NIQE and competitive FID/KID scores, along with the highest HIST value, these results confirm that enforcing spatially continuous statistical fields effectively stabilizes the transport path and reduces structural artifacts introduced by patch-wise translation. Qualitative results in Fig.~\ref{fig3} further demonstrate that SC-UNSB effectively mitigates tiling artifacts and boundary inconsistency, producing spatially coherent cell structures that better support downstream detection. Fig.~\ref{fig4} provides a quantitative and structural comparison of the estimated statistical fields. The radar chart shows that SC-UNSB achieves consistent improvements across multiple quality metrics rather than optimizing a single criterion. The visualization of pixel-wise statistics reveals that the baseline exhibits abrupt transitions at patch boundaries, leading to fragmented normalization patterns. In contrast, SC-UNSB produces smooth and spatially coherent moment fields. Such continuity preserves cellular morphology and staining consistency, which is critical for downstream detection. Additional hyperparameter analyses are reported in Table~\ref{tab:granularity_ablation}.

\begin{table}[ht]
\centering
\footnotesize
\setlength{\tabcolsep}{2.5pt}
\caption{Ablation study of different interpolating granularities.}
\label{tab:granularity_ablation}
\begin{tabular}{l|c|ccc|ccc}
\toprule
\multirow{2}{*}{\makecell[l]{\textbf{Interpolating} \\ \textbf{Granularity}}} & \multirow{2}{*}{\textbf{NIQE}$\downarrow$} & \multicolumn{3}{c|}{\textbf{SC-UNSB (testing set)}} & \multicolumn{3}{c}{\textbf{SC-UNSB (training set)}} \\
\cmidrule(lr){3-5} \cmidrule(lr){6-8}
 &  & \textbf{FID}$\downarrow$ & \textbf{KID}$\downarrow \times 100$ & \textbf{HIST}$\uparrow$ & \textbf{FID}$\downarrow$ & \textbf{KID}$\downarrow \times 100$ & \textbf{HIST}$\uparrow$ \\
\midrule
1 pixel    & 11.38           & \textbf{135.31}  & \textbf{1.807}  & \textbf{0.754}  & \textbf{66.291}  & \textbf{1.534}  & \textbf{0.711} \\
16 pixels  & 11.59           & 147.93           & 2.22            & 0.717           & 70.77            & 1.8             & 0.700           \\
32 pixels  & 11.62           & 162.59           & 3.630           & 0.713           & 77.1             & 2.37            & 0.698           \\
64 pixels  & \textbf{11.34}  & 148.29           & 2.91            & 0.708           & 76.87            & 2.6             & 0.688           \\
128 pixels & 11.40           & 143.37           & 2.47            & 0.704           & 70.6             & 2.07            & 0.682           \\
256 pixels & 13.94           & 143.43           & 2.411           & 0.701           & 68.232           & 1.623           & 0.707           \\
\bottomrule
\end{tabular}
\end{table}

{\textbf{Detection:}} As shown in Table~\ref{tab:1}, training RetinaNet on the ($\mathcal{D}_i$) improves cross-domain detection compared with direct source training. Incorporating LFA or CFA further enhances performance, and their combination achieves the best results, reaching 26.9$\%$ mAP and 45.8$\%$ mAP50 with SC-UNSB. The improvement is most evident when using SC-UNSB-generated data, suggesting that spatially coherent synthesis facilitates more effective feature alignment. Table~\ref{tab:distillation_study} shows that the proposed alignment strategy outperforms conventional distillation methods, which focus on prediction-level transfer without explicitly addressing cross-domain feature mismatch. Additional hyperparameter analyses are reported in Table~\ref{tab:distillation_study} as well.

\begin{table}[ht]
\centering
\footnotesize
\setlength{\tabcolsep}{3.5pt}
\caption{Comparison with different Knowledge Distillation methods and ablation study of Hyperparameter of ours.}
\label{tab:distillation_study}
\begin{tabular}{l|cccc|cccc}
\toprule
\multirow{2}{*}{\textbf{Metric}} & \multicolumn{4}{c|}{\textbf{Knowledge Distillation}} & \multicolumn{4}{c}{\textbf{Hyperparameter Analysis}} \\
\cmidrule(lr){2-5} \cmidrule(lr){6-9}
 & KD\cite{hinton2015distilling} & DKD\cite{zhao2022decoupled} & SPD\cite{wei2024scaled} & \textbf{Ours} & $\alpha=0.1$ & $\alpha=0.2$ & $\alpha=0.3$ & $\alpha=0.4$ \\
\midrule
mAP $\uparrow$   & 21.7\% & 18.6\% & 22.3\% & \textbf{26.9\%} & 25.3\% & \textbf{26.9\%} & 26.1\% & 24.5\% \\
mAP50 $\uparrow$ & 40.6\% & 36.7\% & 43.4\% & \textbf{45.8\%} & 44.2\% & \textbf{45.8\%} & 45.4\% & 45.1\% \\
\bottomrule
\end{tabular}
\end{table}

\section{Conclusion}
This work presents a two-stage framework for cross-domain cervical cell detection that explicitly addresses both appearance-level domain shift and representation-level feature misalignment. By constructing a spatially coherent intermediate domain through SC-UNSB and introducing dual-level feature alignment within a distillation framework, the proposed approach enhances the transfer of domain-invariant knowledge across institutions. These results highlight the potential of combining generative domain bridging with progressive feature alignment to enable cross-domain diagnosis in cervical cytopathology.

\begin{credits}
\subsubsection*{Acknowledgments}
This work was supported by the Natural Science Foundation of Jiangsu Province (BK20251838) and the Nantong Science and Technology Program Project (JC2024055).

\subsubsection*{Disclosure of Interests}
The authors have no competing interests to declare that are relevant to the content of this article.
\end{credits}

\end{document}